%% file: main.tex
\documentclass[lettersize,journal]{IEEEtran}
\usepackage{amsmath,amsfonts}
\usepackage{algorithmic}
\usepackage{algorithm}
\usepackage{array}
\usepackage[caption=false,font=normalsize,labelfont=sf,textfont=sf]{subfig}
\usepackage{textcomp}
\usepackage{stfloats}
\usepackage{url}
\usepackage{verbatim}
\usepackage{graphicx}
\usepackage{cite}
\usepackage[square,sort,comma,numbers]{natbib}
\usepackage{svg}
\usepackage[utf8]{inputenc} 
\usepackage[T1]{fontenc}    
\usepackage{hyperref}       
\usepackage{url}            
\usepackage{booktabs}       
\usepackage{amsfonts}       
\usepackage{nicefrac}       
\usepackage{microtype}      
\usepackage{xcolor}         
\usepackage{graphicx} 
\usepackage{multirow}
\usepackage{svg} 
\usepackage{makecell}
\usepackage{pifont}
\usepackage{wrapfig}
\usepackage{enumitem}
\usepackage{diagbox}
\usepackage{amsmath}
\usepackage{graphicx}
\usepackage{array}
\usepackage{caption}
\usepackage{orcidlink}
\title{FineFake: A Knowledge-Enriched Dataset for Fine-Grained Multi-Domain Fake News Detection}

\author{Ziyi Zhou, Xiaoming Zhang,  Litian Zhang, Jiacheng Liu, Senzhang Wang, \\Zheng Liu, Xi Zhang, Chaozhuo Li and Philip S. Yu \orcidlink{0000-0002-3491-5968}, \textit{Fellow, IEEE}
\thanks{

Ziyi Zhou, Xiaoming Zhang, Litian Zhang and Jiacheng Liu are with School of Cyber Science and Technology, Beihang University, Beijing 100191, P. R. China (e-mail: ziyizhou@buaa.edu.cn; yolixs@buaa.edu.cn; litianzhang@buaa.edu.cn;liujc11@buaa.edu.cn). (Corresponding author: Xiaoming Zhang.)

Senzhang Wang is with the School of Computer Science and Engineering,
Central South University, Changsha 410083, China. (e-mail: szwang@csu.edu.cn).

Zheng Liu is with Beijing Academy of Artificial Intellligence.

Xi Zhang is with School of Cyber Science and Technology, Beijing University of Posts and Telecommunications,
Beijing 100876, (e-mail: zhangx@bupt.edu.cn).

Chaozhuo Li is with School of Cyber Science and Technology, Beijing University of Posts and Telecommunications,
Beijing 100876, (e-mail: lichaozhuo@bupt.edu.cn).

Philip S. Yu is with the Department of Computer Science, University of Illinois at Chicago, Chicago, IL 60607 USA (e-mail: psyu@uic.edu).
}

}

\begin{document}
\maketitle


\begin{abstract}

Existing benchmarks for fake news detection have significantly contributed to the advancement of models in assessing the authenticity of news content. However, these benchmarks typically focus solely on news pertaining to a single semantic topic or originating from a single platform, thereby failing to capture the diversity of multi-domain news in real scenarios. 
In order to understand fake news across various domains, the external knowledge and fine-grained annotations are indispensable to provide precise evidence and uncover the diverse underlying strategies for fabrication, which are also ignored by existing benchmarks. 
To address this gap, we introduce a novel multi-domain knowledge-enhanced benchmark with fine-grained annotations, named \textbf{FineFake}.
FineFake encompasses 16,909 data samples spanning six semantic topics and eight platforms. Each news item is enriched with multi-modal content, potential social context, semi-manually verified common knowledge, and fine-grained annotations that surpass conventional binary labels. Furthermore, we formulate three challenging tasks based on FineFake and propose a knowledge-enhanced domain adaptation network.  
Extensive experiments are conducted on FineFake under various scenarios, providing accurate and reliable benchmarks for future endeavors. 
The entire FineFake project is publicly accessible as an open-source repository at \url{https://github.com/Accuser907/FineFake}.
\end{abstract}
\maketitle
\begin{IEEEkeywords}
Fake News Detection, Multi-Domain Benchmark, Fine-Grained Classification.
\end{IEEEkeywords}

\section{Introduction}
\label{introduction}
In the contemporary landscape of the ever-evolving digital society, social media stands as a prominent  medium for accessing news. 
It has emerged as an optimal milieu for the dissemination of falsified information, posing a significant threat to both individuals and society~\cite{zhou2020survey}.
For example, during the COVID-19 infodemic, the spread of fake news caused incorrect medical interventions, leading to social unrest and numerous fatalities~\cite{rocha2021impact,greene2021quantifying}.
Hence, automatic fake news detection has become crucial, drawing significant academic focus. 
To enhance the pursuit of identifying fake news, a series of datasets like Twitter~\cite{boididou2015verifying} and Pheme~\cite{zubiaga2017exploiting} has been developed, evolving from small, unimodal sets to large, multimodal compilations. These enhancements have  augmented the wealth of extensive and diverse information available for further analysis.

\input{Table/MVAE_ex_1}
Notwithstanding the notable advancements in fake news detection datasets, existing datasets are generally constructed upon the news centered around a similar topic or a single platform, leading to the limited generality. 
For instance, as illustrated in Table~\ref{tab:addlabel}, the LIAR and Breaking datasets \cite{wang2017liar,pathak2019breaking} only comprise samples related to the topic of politics, whereas the Weibo and Twitter datasets \cite{jin2017multimodal,boididou2015verifying} solely consist of news sourced from a single platform. 
Nevertheless, within the sphere of different real-world news platforms, an incessant deluge of millions of news articles spanning various topics. 
The diverse range of topics and platforms is largely ignored by existing datasets, leading to inadequate assessments of cross-domain capability. 
For example, as depicted in Table~\ref{mvae_ex_1}, when a widely-used detection model MVAE \cite{khattar2019mvae} is trained on a specific news topic or platform and then applied to another topic or platform, its performance exhibits a notable decline. 
The underlying reasons may be attributed to two key aspects. 
From the perspective of semantic topics, tokens commonly associated with fake news within certain domains, such as ``vaccine'', ``virus'', and ``side-effect'' in the health domain, may be notably scarce in other domains like business. Consequently, the semantic distributions of news across topics are apt to diverge significantly, introducing the classical covariate shift problem \cite{shimodaira2000improving}. 
From the perspective of platforms, the proportion of fake news can vary significantly across different platforms. For instance, reputable sources like CNN generally feature higher credibility compared to self-media posts on Twitter, which may have a higher proportion of true news~\cite{bovet2019influence}. This imbalance in the proportion of real and fake news across platforms introduces another classic challenge in domain adaptation: the label shift problem \cite{garg2020unified}. 
These two domain shifts underscore the necessity for a benchmark dataset for fake news to evaluate a model's domain adaptation capabilities.

\begin{figure}[!t]
    \centering
    \includegraphics[width=1\linewidth]{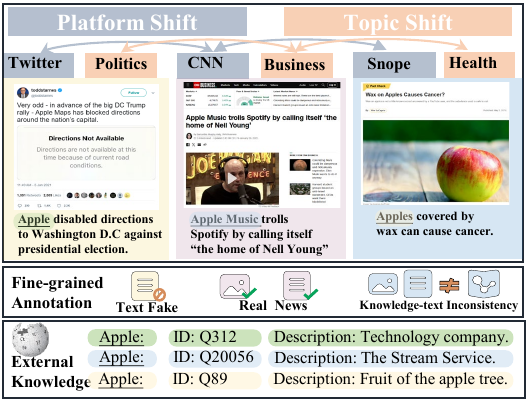} 
    \caption{The proposed FineFake: a multi-domain dataset that encompasses instances from diverse platforms and topics. Each sample is associated with corresponding image, accurate knowledge and fine-grained label.}
    \label{fig:f1}
\end{figure}
To alleviate the aforementioned challenges of domain shift, external knowledge graphs like ConceptNet~\cite{speer2017conceptnet} are incorporated to provide extra cross-domain information, serving as a bridge between disparate domains \cite{dun2021kan,chen2022evidencenet,hu2021compare}. 
The determination of the veracity of news often relies on foundational knowledge and latest external information as auxiliary factors for assessment. 
However, the same entity may have different meanings across domains, causing entity ambiguity. As depicted in Fig~\ref{fig:f1}, the term ``apple'' within the health domain refers to a fruit, whereas in the business domain, it denotes the company Apple Inc. Consequently, these disparities across domains give rise to the challenge of entity ambiguity, which can introduce noisy knowledge and degrade the performance of the model.
Therefore, the development of datasets containing accurate  common knowledge is essential for advancing fake news detection.

\input{Table/table_dataset}
Another significant aspect of fake news across different topics or platforms lies in the diverse underlying strategies used for fabrication. 
For instance, Twitter has a more proportion of news with fake images than CNN, which can be attributed to the higher prevalence of manipulated images on social media platforms compared to official news sources~\cite{nygren2018changing,gupta2013faking}. Therefore, it is essential to identify distinguishing features of fake news that can provide a rational basis for judgment across platforms. 
However, conventional fake news detection datasets typically classify news articles into binary classes (real or fake) or employ broad categories such as ``most likely'' or ``somewhat likely''~\cite{wang2017liar,yang2024towards}. Unfortunately, such coarse-grained annotations fail to reveal the factors contributing to the falseness of a news item, such as fabricated images or inconsistencies between text and image. 
Therefore, there's an urgent need for a fine-grained annotation strategy to uncover the reasons behind fake news.

In order to address the aforementioned challenges, in this paper, we propose a comprehensive and knowledge-enhanced dataset for fake news detection, dubbed FineFake. 
As illustrated in Table~\ref{tab:addlabel}, FineFake surpasses its predecessors by spanning multiple topics and platforms, enjoying accurate common knowledge and fine-grained annotations, thereby furnishing robust and solid data support for further research endeavors. 
FineFake encompasses instances collected from diverse media platforms, such as CNN, Reddit and Snopes. All instances are also labeled into six distinct topics (i.e., politics, entertainment, business, health, society and conflict). 
The inclusion of this comprehensive multi-domain dataset fosters a deeper comprehension of correlations between fake news from various domains. 
Each news article contains textual content, images, possible social connections, and other pertinent meta-data. 
To ensure the provision of reliable common knowledge, each news is appended with the relevant knowledge entities and descriptions in a semi-manual labeling manner. 
Moreover, we introduce an innovative annotation guideline that extends beyond traditional binary class labels. Our approach incorporates a six-category annotation strategy that sheds light on the reasons behind the detected fake news, including real,  textual fake, visual fake, text-image inconsistency, content-knowledge inconsistency, and other samples. 
Based on FineFake, we conduct extensive experiments on data characteristics, fine-grained classification and domain adaptation, establishing a valuable benchmark for future research.
Furthermore, to solve the  covariate and label shift issues in FineFake, we propose a \textbf{k}nowledge-\textbf{e}nhanced domain \textbf{a}daptation \textbf{n}etwork, dubbed \textbf{KEAN}, which achieves SOTA performance in most scenarios.
Our contributions are summarized as: 
\begin{itemize}[leftmargin=0.2cm, itemindent=0.2cm]
\item FineFake dataset represents a pioneering effort for cross-domain fake news detection, which systematically gathers and formalizes multi-modal news content from diverse topics and platforms. 
\item The FineFake dataset enhances each news by incorporating rich and reliable  external knowledge through semi-manual labeling, which ensures the provision of accurate evidence.
\item Different from the conventional binary class-based annotations, FineFake employs a fine-grained labeling scheme that classifies news articles into six distinct categories, elucidating the underlying reasons behind the formation of fake news. 
\item We propose \textbf{KEAN}, a \textbf{k}nowledge-\textbf{e}nhanced domain \textbf{a}daptation \textbf{n}etwork model for fake news detection. We conduct extensive experiments to evaluate the performance of SOTA approaches on FineFake, furnishing valuable benchmarks and shedding light on avenues for future research.
\end{itemize}

\section{Related Work}
\subsection{Fake News Detection Datasets}
Since the rapid development of the internet, a proliferation of publicly available datasets concerning the detection of fake news has ensued. Initially, researchers predominantly focused on collecting textual data, concentrating on specific domains to construct these datasets~\cite{wang2017liar,thorne2018fever,pathak2019breaking,patwa2021fighting}. 
For instance, the LIAR dataset~\cite{wang2017liar} harnesses Politifact\footnote{http://www.politifact.com/} to extract news items, subsequently annotating them with six classification labels. 
Similarly, FEVER~\cite{thorne2018fever} is generated by altering sentences extracted from Wikipedia and pre-processing Wikipedia data, each claim is annotated to a three-way classification label. Additionally, the COVID-19 dataset~\cite{patwa2021fighting} meticulously curates a manually annotated repository of social media posts and articles pertaining to COVID-19, aiming to facilitate research endeavors in identifying pertinent rumors that possess the potential to instigate significant harm.

The aforementioned uni-modal datasets primarily concentrate on linguistic analysis, thus disregarding the crucial dimensions of social networks for dissemination, corresponding images, and metadata, essential for a more comprehensive detection framework. In stark contrast to traditional textual news media releases, the presence of multimodal news incorporating images or videos tends to captivate greater attention and propagate more extensively, thus may lead to more damage transmission. Consequently, there has been a discernible surge in the construction of datasets integrating images and social network data~\cite{shu2020fakenewsnet,zhang2018fauxbuster,nielsen2022mumin,zubiaga2017exploiting,boididou2015verifying}. 
For instance, MM-COVID~\cite{li2020mm} offers a multilingual dataset encompassing news articles augmented with pertinent social context and images, aimed at facilitating the detection and mitigation of fake news pertaining to the COVID-19 pandemic. Similarly,  Weibo~\cite{jin2017multimodal} collects original tweet texts, attached images, and contextual information sourced from Weibo, a prominent Chinese microblogging platform renowned for its objective ground-truth labels.

Although the existing multi-modal datasets contain multi-modal data like images, the majority of them overlook critical factors such as multi-domain attributions, external knowledge and fine-grained classification annotations. $MR^2$~\cite{hu2023mr2} attempts to address this gap by incorporating evidence retrieved from online sources as metadata to enhance fake news detection. Nonetheless, this approach of online retrieval lacks a guarantee of the accuracy of external evidence, potentially introducing extraneous noise information instead.
Additionally, Weibo21 ~\cite{nan2021mdfend} divides textual news data into nine distinct topics, yet it only focus on topics, overlooking the the multitude of platforms through which news dissemination occurs. Although LIAR~\cite{wang2017liar} and RD-E~\cite{yang2024towards} categorize fake news into six categories, they rely on pre-existing labels from fact-checking websites such as ``most likely'' or ``somewhat likely'', failing to explore the underlying reasons behind the identification of fake news.

\subsection{Fake News Detection Methods}
At the outset, a considerable amount of research focused on refining the extraction of semantic features inherent in news content itself, recognizing the wealth of information embedded within the content conducive to discerning its veracity~\cite{ma2016detecting,khattar2019mvae,Zhou2020SAFESM,chen2022cross}. However, the escalating convergence of semantic structures between fake and authentic news has rendered the task of distinguishing between them based solely on semantics increasingly formidable~\cite{zhou2020survey}.  Consequently, attention has shifted towards leveraging external knowledge as supplementary information to bolster fake news detection efforts~\cite{qian2021knowledge,dun2021kan,sun2023inconsistent,sun2021inconsistency,hu2021compare}. For instance, CompareNet~\cite{hu2021compare} constructs a directed heterogeneous document graph to compare news to external knowledge base through the extractied entities. 

Nonetheless, news content exhibits significant variations across diverse platforms and topics, and the distribution of fake and authentic news also fluctuates accordingly~\cite{zhou2020survey,zhu2022memory}. Effective deployment of a well-trained fake news detection model in real-world scenarios necessitates robust cross-domain capabilities. However, only a limited number of studies have earnestly tackled the challenges posed by multi-domain and cross-domain fake news detection~\cite{wang2018eann,nan2021mdfend,zhu2022memory,yue2022contrastive}. Consequently, we construct a multi-domain knowledge-enhanced multimodal fine-grained dataset that holds immense potential for facilitating research in the realms of multi-domain and cross-domain fake news detection in reality.

\section{FineFake: The Knowledge-Enriched Fine-Grained Multi-Domain Dataset} 

\input{Table/snopes_topic}
\subsection{Multi-domain News Collection} 
    
\begin{figure}[!t]
    \setlength{\abovecaptionskip}{0.15cm}
    \begin{center}
    \includegraphics[scale=0.8]{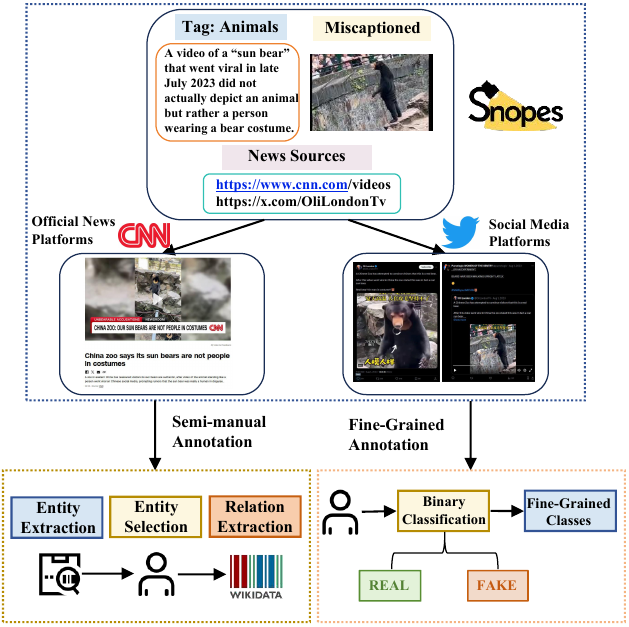}
    \end{center}
    \caption{The Construction Process of FineFake. Snopes is used as the starting point for data collection and the external links within the claim explanations are leveraged as sources for multi-platform data collection. Platforms are categorized into official news platforms and social media platforms while FineFake also collects potential social network information from social media platforms. Finally, each piece of news undergoes semi-manual knowledge annotation and fine-grained annotation to ensure label accuracy.}
    \label{fig:construction}
\end{figure}

Current fake news detection datasets predominantly gather data from singular platforms within narrow topics~\cite{zhou2020survey}. To develop a large-scale dataset spanning multiple platforms and topics with substantial diversity, we employ a comprehensive data collection strategy involving three primary channels: Snopes, social media platforms (e.g., Twitter), and official news websites (e.g., CNN). The overall collection process for FineFake is illustrated in Figure~\ref{fig:construction}.

Snopes~\footnote{https://www.snopes.com} is a website that verifies the authenticity of news reports. Each verified claim encompasses a substantial amount of additional information regarding the original news post, including external links to the original source, expert-assigned topic labels, and authenticity assessments provided by professionals. 
The provision of links for news provenance can enrich the sources of news across multiple platforms, while the tagging of topics facilitates the construction of multi-topic databases. 
Moreover, the  authenticity ratings provided by professionals are advantageous for our exploration of the underlying causes of fake news.
Unlike previous approaches that only use the summarized claims as samples~\cite{hu2023mr2}, we collect data on various topics through the topic categorization provided by Snopes, preserving elements such as claims, images, tags, verification materials and rating categories. 
The external news links are then used as springboards to gather data from multiple platforms for multi-source news collection.
For official news websites, APNews, CNN, New York Times, The Washington Post, and the CDC are chosen due to their credibility and the high quality of content. 
The tailored web crawlers are programmed to meticulously extract not just the textual content but also images, authors, publication dates and other relevant metadata from each article. 
In parallel, we extend our data collection framework to integrate social interactions from social media platforms, including Twitter and Reddit, which contain abundant social information that is invaluable for understanding how online news spreads and evolves. For Twitter, the well-documented API is utilized to collect tweets, retweets, replies of the news links found on Snopes. 
To effectively gather data from Reddit, a specialized crawler is designed to obtain posts and user interactions due to its lack of API, enabling us to build another layer of social network reflecting the news events.

Since data in Snopes is affiliated with tags, to facilitate the study of distinctions among different topics,  these tags are  consolidated into six categories within the Snopes dataset. 
The six topic categories, including politics, entertainment, business, health, society and conflict are carefully chosen based on their prevalence and relevance in both Snopes tags and fake news detection~\cite{liu2021dtn,nan2021mdfend}. The specific meanings of thses six topics are described as follows:
\begin{itemize}[leftmargin=0.2cm,itemindent=0.2cm]
\item \textbf{Politics.} The politics topic records the statements made by many politicians during elections and political activities, along with corresponding news photographs.
\item \textbf{Entertainment.} The entertainment topic focuses on hot news events and related reports about celebrities in the entertainment industry.
\item \textbf{Business.} The business topic represents news about the business plans and actions of entrepreneurs and well-known companies.
\item \textbf{Health.} The health topic aims at news related to major health events like COVID-19, fake news often leads to social panic and the spread of incorrect treatment methods.
\item \textbf{Society.} The society topic includes news about society events, such as education reform, economic demelopment, residents' quality of life and so on.
\item \textbf{Conflict.} The conflict topic focuses on news about wars, military and conflict between countries and regions.
\end{itemize}
The correlation between the category labels in Snopes and our designated topic labels is elucidated in Table~\ref{snope}.
News collected from the other two platforms are annotated with the same topic as the source data in Snopes.

Given our overarching objective of constructing a multimodal dataset and benchmark exclusively in English, text-only posts and news articles published in languages other than English are filtered. Subsequently,  data deduplication techniques is implemented to eradicate redundant news items, thereby mitigating the risk of data leakage. Moreover, to uphold data quality standards, we eliminate instances containing unqualified images or excessively brief textual content. 

Ultimately, instances are collected from three different types of platforms among six topics by using Snopes as the base, ensuring the diversity of FineFake. 
Each data point includes textual content, corresponding images, metadata, and any associated social media presence. The integration of multiple platforms and diverse topics underscores the multi-domain characteristics of FineFake, laying the groundwork for subsequent cross-platform and cross-topic research.


\subsection{Semi-manual Knowledge Alignment}  
One notable feature of FineFake is its incorporation of pertinent background knowledge, which complements the original multi-modal content and holds immense potential to advance the identification of fake news. 
A semi-manual knowledge alignment strategy is adopted to accomplish the integration. 
Specifically, an entity link tool ~\cite{ferragina2010tagme} is employed to adeptly recognize named entities within the news text. 
To ensure the accuracy of extracted named entities, we initially set a lower threshold $\omega_1$ to compile the initial list of entities and subsequently adjust the threshold higher to $\omega_2$ to identify entities with higher confidence. Entities extracted within the threshold range of $\omega_1$ to $\omega_2$ are verified by human annotators for contextual accuracy, achieving disambiguation.
Subsequently,  all triplets within Wikidata with a distance of one from the entity are retrieved, including relationship-type triplets  and attribute triplets.
Relationship-type triplets denotes as $(h,r,t)$, where $h$ denotes the head entity, $r$ denotes the relation and $t$ denotes the tail entity. They represent entities connected to the extracted entities in Wikipedia, these triplets are widely utilized by knowledge-enhanced methods to capture the background knowledge of news~\cite{zhang2024reinforced,hu2021compare}. Attribute triplets denotes as $(e,r_d,d)$, where $e$ denotes the entity, $r_d$ denotes the relation is description of the entity and $d$ denotes the text of description.
This augmentation adds knowledge assistance, thereby facilitating more nuanced analyses and interpretations of fake news instances.

\subsection{Fine-grained Human Annotation}
\label{fine-grained section}

Diverging from traditional binary categorization schemes, the FineFake dataset further introduces a novel classification framework wherein each instance of fake news is assigned to one of six distinct categories, namely real, text-based fake, image-based fake,  text-image inconsistency, content-knowledge inconsistency and others. These categories are determined based on the underlying reasons that contribute to the falseness of the news, thereby providing a more nuanced understanding of the deceptive nature of the content.

The first two categories, namely text-based fake and image-based fake, denote instances where the falsity of the news can primarily be discerned through analysis of either the textual content or the accompanying images.  
Text-image inconsistency represents a category wherein the fake news is classified as such due to the evident disparities and contradictions between the textual content and the associated images. 
The fourth category, content-knowledge inconsistency, encompasses cases where the news content, including both textual content and images, contradicts externally retrieved knowledge. 
Lastly, the ``others'' category encompasses instances that do not fall squarely within the aforementioned categories but still exhibit deceptive characteristics. This category ensures the inclusion of diverse and anomalous cases, capturing a broad spectrum of fake news manifestations that may not fit neatly into the predefined categories. 
 
Therefore, 
in the verification stage, 5 professional annotators are engaged to annotate the binary classification label and fine-grained label for each data with the usage of Snopes as references.  
To promote consistency and mitigate subjectivity, we implement inter-annotator agreement measures. Each news is independently labeled by five annotators to assess consistency and calculate agreement scores (e.g. Fleiss's kappa). Discrepancies are resolved through discussions among annotators.
The refined categorization scheme in FineFake facilitates a deeper analysis of the underlying reasons of fake news, empowering researchers to develop sophisticated detection methods that account for distinct modalities, inconsistencies, and the intersection between textual and visual elements.

\subsection{Fine-Grained and Multi-Domain Fake News Detection Tasks}
\label{section:3.2}
Based on the constructed FineFake dataset, we propose three downstream tasks to evaluate the performance of SOTA fake news detection models under various scenarios. 

\subsubsection{Binary classification task}
The primary objective in fake news detection models is to classify news articles as either fake or true. 
To investigate the efficacy of external knowledge in enhancing fake news detection, we conduct two types of binary classification tasks: one without any external knowledge and another with knowledge augmentation. 
\begin{equation}
\label{equation1}
\mathcal{L}(y,\hat{y}) = -y \log(\hat{y})- (1-y) \log(1-\hat{y}), 
\end{equation}
in which $y$ denotes the actual label and $\hat{y}$ denotes the probability of model's output. 
\begin{figure*}[t]
\centering
\includegraphics[scale=0.95]{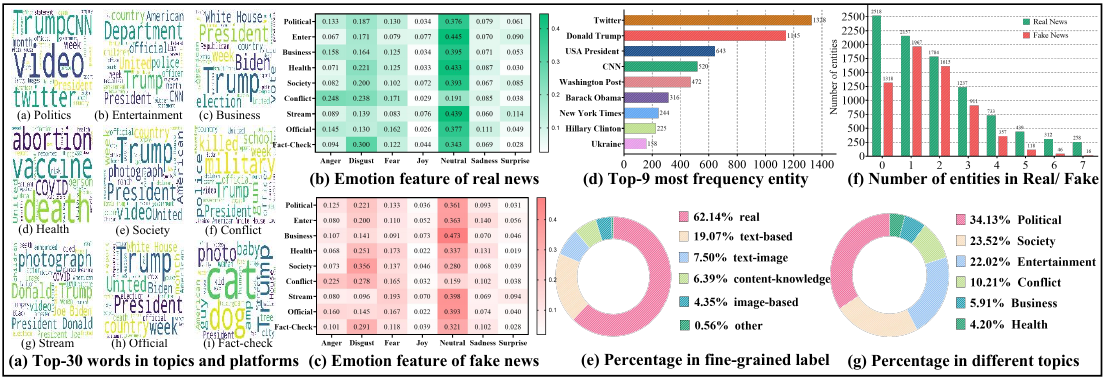}
    \caption{Basic information and statistic analysis on FineFake.}
    \label{fig:statistic}
\end{figure*}

\subsubsection{Fine-grained classification task}
Currently, both fake news detection models and datasets predominantly concentrate on accurately predicting the authenticity of news articles, often overlooking the determination of the reasons behind fake news. To address this limitation, a fine-grained classification task is proposed to expand traditional binary classes into six classes, as detailed defined in Section~\ref{fine-grained section}. In the training process, we utilize cross-entropy loss as loss function:

\begin{equation}
\label{equation2}
\mathcal{L}(y,\hat{y}) = -\sum_{i=1}^{N} y_i \log(\hat{y}_i), 
\end{equation}
where $N$ denotes the number of label categories. 

\subsubsection{Multi-domain adaptation task}
To evaluate cross-domain capacity of existing models, we design three cross-domain tasks: topic adaptation, platform adaptation and dual domain adaptation.
Topic adaptation is proposed to measure the model's capability to overcome data distribution variance  between different topics from the same platform. In this task, models are trained on four topics and tested on the remaining two topics. Platform adaptation, on the contrary, aims to test the model's ability to adapt to label shift problem, with training on one platform and testing on another under the same topic.
The most challenging task, dual domain adaptation, requires models to simultaneously adapting across both topics and platforms.
This task presents a significant challenge, as it requires the model to exhibit generalization abilities enabling it to perform well when confronted with data from previously unseen domains and deviated label distribution.
\input{Table/dataset_analysis}
\section{Comprehensive Data Analysis of FineFake}

In this section, we analyze the FineFake dataset to understand its complex structure and characteristics. 
Table~\ref{tab:dataset_analysis} shows detailed statistics, including the number of true and fake samples in each topic/platform, the average text length, and average entities for each news.  
The obvious differences in text length across three categories of platforms ensures the diversity of data. Furthermore, the imbalanced distribution of positive and negative samples in the three platforms are also significant. Official news sources generally exhibit higher credibility, thus having a higher proportion of true news. In contrast, Snopes has a higher proportion of fake news as it focuses on debunking fake news events. 

\textbf{(a) Multi-Domain Analysis.}
Figure \ref{fig:statistic}(a) presents the top-30 words observed in the six topics and three platforms. 
Notably, one can easily observe that high-frequency vocabulary in different domains exhibits distinct patterns and thematic clusters. 
Such findings shed light on the domain-specific linguistic variations, revealing the importance of constructing a multi-domain fake news dataset.
\textbf{(b) Emotion Tendency.} Figure~\ref{fig:statistic}(b) and (c)  illustrate the average value of emotion tendencies of the nine domains, which are calculated by Emotion DistilRoBERTa~\cite{hartmann2022emotionenglish}. 
The emotion distributions of various domains are apparently different, further demonstrating the value of FineFake dataset. 
One can also see that the emotion tendencies of true/fake news are also different within the same domain, such as the ``society'' topic exhibits a more prominent ``disgust'' emotion in fake news than the real ones. In ''conflict'' topic, there is a significant decrease in neutral sentiment compared to other topics, replaced by a predominant sense of anger.  
This finding is aligned with previous literature \cite{alonso2021sentiment,bhutani2019fake} that sentiments also contribute to advancing the performance of fake news detection.  
\textbf{(c) External Knowledge Analysis.} Figure~\ref{fig:statistic}(d) and (f) analyze the presence of external knowledge entities. The results reveal that approximately 85\% of the news articles encompass external knowledge entities, with real news articles demonstrating a higher propensity to incorporate such entities compared to fake news articles.  Figure \ref{fig:statistic}(d) illustrates the top-9 most frequently extracted entities and the top-3 entities are ``Twitter'', ``Donald Trump'' and ``USA President''.
\textbf{(d) Data Proportion Analysis.}
Figure \ref{fig:statistic}(e) provides an overview of the proportions of data with fine-grained labels. This demonstrates FineFake's exploration into the fundamental causes of fake news, thus providing a reliable benchmark for fine-grained analysis in future work. Moreover, the proportions of ``text-image inconsistency'' and ``content-knowledge inconsistency'' within the fake news category also highlight the importance of multimodal information and external knowledge.
Figure \ref{fig:statistic}(g) provides an overview of the proportions of data from each topic domain in relation to the total dataset, thereby highlighting the multi-domain characteristic of our study.



\begin{figure}[!t]
    \begin{center}
    \includegraphics[scale=0.96]{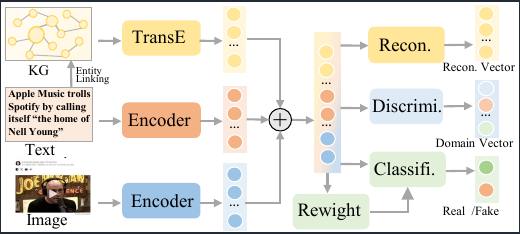}    \end{center}    \caption{The overview of KEAN model. }
    \label{fig:model}
    \vspace{-0.2cm}
\end{figure}
\section{KEAN: The Knowledge-Enhanced Domain Adaptation Network}
To address both the covariate shift and label shift problem simultaneously, we propose a \textbf{k}nowledge-\textbf{e}nhanced domain \textbf{a}daptation \textbf{n}etwork, dubbed \textbf{KEAN}. As Fig~\ref{fig:model} illustrates, the structure of KEAN is based on the architecture of Domain Adversarial Neural Networks (DANN)~\cite{ganin2016domain}.
As knowledge graph contains rich domain commonsense knowledge~\cite{chen2023knowledge,zeng2023knowledge,ghosal2020kingdom},
three encoders are utilized for text, visual and knowledge modeling, respectively.
Furthermore, inspired by previous work~\cite{garg2023rlsbench,li2019target,tachet2020domain},
a reconstruction module and a reweight module is implemented to help solve the covariate shift and label shift simultaneously.

\noindent{\textbf{Multimodal Encoder.}}
An instance is defined as a tuple $I = \{T,V,E_T^n\}$ representing three modalities of contents: the textual content $T$ and the visual content $V$ of the news, and a set of textual entities $E_T^n$, where $n$ represents the number of entities. Additionally, the constructed knowledge graph is defined as $KG$. 
A pre-trained CLIP~\cite{Radford2021LearningTV} is utilized as the encoder in our method, which transforms sentences and images into embeddings through its robust multimodal representation capabilities:

\begin{gather}
\label{equation3}
\begin{aligned}
h_t &= \mathrm{CLIP}_{text}(T), h_t \in \mathcal{R}^{d1} \\
h_v &= \mathrm{CLIP}_{visual}(V), h_v \in \mathcal{R}^{d2} \\
\end{aligned}
\end{gather}
\noindent{\textbf{Knowledge Graph Encoder.}} As each instance in FineFake contains external knowledge from wikidata, the one-hop neighbours of the entities $E_T^n$ are extracted to construct a sub-graph of $KG$ by aggregating all the triplets, defined as $KG_{sub}$. Specifically, TransE~\cite{bordes2013translating} is utilized as our knowledge graph embedding method due to its simplicity and effectiveness. 
After feature extraction, each node $j \in KG_{sub}$ has its representation $h_j$. Then the average of the feature vectors $h_j$ for all nodes in $KG_{sub}$ is computed to obtain the final graph feature $h_{kg}$ as the representation:
\begin{equation}
\label{equation4}
\begin{aligned}
h_{kg} = \frac{1}{|E|}\sum_{E_j}^{E} \mathrm{TransE}(E_j) , E_j \in KG_{sub}
\end{aligned}
\end{equation}
The representations $h_t,h_v,h_{kg}$ are then transformed to $h_t^{'},h_v^{'},h_{kg}^{'}$ through fully connected layers.  The final feature representation is obtained by concatenation: $h_{I}=[h_t^{'};h_v^{'};h_{kg}^{'}]$.

\noindent{\textbf{Domain-adversarial Training.}} Based on the DANN architecture, KEAN comprises a task classifer $C$ (with parameters $\theta_C$), a domain-discriminator $D_{adv}$ (with parameters $\theta_D$) and a decoder $D_{recon}$ (with parameters $\theta_R$). $C$ and $D_{adv}$ focus respectively on news truthfulness and domain differentiation:

\begin{equation}
\label{equation5}
\begin{aligned}
\mathcal{L}_{cls} = E_{I_s} &(-\sum_{i=1}^{N} y_i \log C(h_I)) \\
\mathcal{L}_{adv} = -E_{I_s}(\log D_{adv}&(h_{I_s})) - E_{I_t}(\log(1-D_{adv}(h_{I_t}))\\
\end{aligned}
\end{equation}
To further enforce domain-invariance into the encoded representation $h_{kg}^{'}$, A decoder $D_{recon}$ is utilized with a  reconstruction loss, as proposed in prior work~\cite{bousmalis2016domain}:

\begin{equation}
\label{equation6}
\begin{aligned}
\mathcal{L}_{recon}(I_s,I_t) = E_{h_{kg}}(\left\| D_{recon}(h_{kg}^{'})-h_{kg}\right\|^{2})
\end{aligned}
\end{equation}
The final optimization of the domain-adversarial training is based on the minimax game: where $\alpha$ and $\beta$ are hyper-parameters. The minimax game is realized by reversing the gradients of $\mathcal{L}_{adv}$ while back-propagation with a reverse layer~\cite{ganin2016domain}:
\begin{equation}
\label{loss}
\begin{aligned}
\mathcal{L}_{loss} = (\mathcal{L}_{cls}+ &\alpha \mathcal{L}_{adv}+ \beta \mathcal{L}_{recon}) \\
\hat{\theta_C},\hat{\theta_R} = arg  \underset{\theta_C,\theta_R} {\text{min}}  \mathcal{L}_{loss} ,  &  \hat{\theta_D} = arg  \underset{\theta_D}{\text{max}} \mathcal{L}_{loss}
\end{aligned}
\end{equation}

\noindent{\textbf{Re-weighting.}} Re-weighting the classifier is a widely used technique to address label shift problems~\cite{garg2023rlsbench}. Following the approach outlined in BBSE~\cite{lipton2018detecting}, the classifier is re-weighted by estimating the distribution of labels in target domain. Specifically, We seek to obtain the importance weights $\hat{w}_t(y)$, defined as the ratio of the probability of observing label $y$ in the target domain to that in the source domain, i.e., $\hat{w}_t(y)=\frac{p_t(y)}{p_s(y)}$.
To calculate $\hat{w}_t(y)$, the classifier is trained with source data while the confusion matrix $C_h$  of the source domain and probability mass function $q_h$ of $f(X)$ under predicted target distribution is then calculated. Then, $\hat{w}_t(y)$ can be obtained by $C_h$ and $q_h$ as Equation~\ref{equation7}. As $\hat{w}_t(y)$ is obtained, the classifier is retrained by the importance weighted loss as followed:
\begin{equation}
\label{equation7}
\begin{aligned}
\hat{w}_t(y) = & C_h^{-1}*q_h \\
\mathcal{L}_{loss}^{'} = \frac{1}{n} \sum_{j=1}^{n}& \hat{w} (y_j) \mathcal{L}_{loss} (y_j, f(x_j))
\end{aligned}
\end{equation}
\input{Table/binary_classification}
\section{Experiments}
\subsection{Baseline Models} 
\label{Baseline models}
Following previous works \cite{zhou2020survey}, we select the following fake news detection methods as baselines and divide them into three categories: content-based  method, knowledge-enhanced method and multi-domain method. 

\noindent{\textbf{Content-Based Multimodal Method.}}
\begin{itemize}
\item \textbf{MVAE}~\cite{khattar2019mvae} comprises three components: an encoder to encode the shared representation of features, a decoder to reconstruct the representation, and a detector to classify the truth of posts.

\item \textbf{SAFE}~\cite{Zhou2020SAFESM} calculates the relevance between textual and visual information and defines it as cosine similarity modification to detect fake news. 

\end{itemize}

\noindent{\textbf{Knowledge-Enhanced Method.}}
\begin{itemize}
\item \textbf{CompNet}~\cite{hu2021compare} constructs a directed heterogeneous document graph to utilize knowledge base.
\item \textbf{KAN}~\cite{dun2021kan} incorporates semantic-level and knowledge-level representations in news to improve the performance for fake news detection.
\item \textbf{KDCN}~\cite{sun2023inconsistent}
captures two level inconsistent semantics in one unified framework to detect fake news. 
\end{itemize}

\noindent{\textbf{Multi-Domain Method.}}
\begin{itemize}
\item \textbf{EANN}~\cite{wang2018eann} firstly utilizes a discriminator to derive event-invariant features for multi-domain fake news detection.
\item \textbf{MDFEND}~\cite{nan2021mdfend} introduces a multi-domain fake news detection model that leverages a domain gate to aggregate multiple representations extracted by a mixture of experts.  
\item \textbf{M$^{3}$FEND}~\cite{zhu2022memory} proposes a memory-guided multi-view framework to address the problem of domain shift and domain labeling incompleteness. 
\item \textbf{CANMD}~\cite{yue2022contrastive} proposes a contrastive adaptation network to solve the label shift problem in early misinformation detection.
\end{itemize}

\subsection{Implementation Details}
\label{section:4.2}
All experiments are conducted using the proposed FineFake dataset. 
In classification experiments, instances within FineFake  are split into the ratio of 6:2:2 for training, evaluating and testing, respectively. In domain adaptation experiments, source domain data are split into the ratio of 9:1 for training and evaluating, while all data from target domain are used for testing. 
The experiments are executed on a computational setup consisting of 4 NVIDIA 3090 GPUs, each equipped with 24GB of memory. We set $\alpha$ as 0.8 and $\beta$ as 0.4 for loss function.
For optimization purposes, we employ the AdamW optimizer \cite{loshchilov2017decoupled} with a weight decay value of 5e-4. The batch size is set to 32, and the initial learning rate is established at 1e-3. Subsequently, the learning rate is decayed gradually with each epoch. In order to eliminate the potential impact of random variations,  the random seed is fixed throughout the experiments.
The hyperparameters of all baseline models are carefully tuned on the validation sets to achieve an optimal configuration. 
\subsection{Experimental Results}
\subsubsection{Binary Classification Task}
In this study, we perform binary classification analysis under two conditions: knowledge enhancement and no knowledge. Experimental results are presented in Table \ref{tab:binary-classification}. Notably, 
KDCN and KEAN exhibit the highest improvement when trained on the knowledge-augmented dataset, indicating their robust capacity to assimilate and leverage external knowledge. 
Across all knowledge-enhanced models, training with knowledge enhancement yields a substantial improvement in all four metrics. 
This finding demonstrates the significance of incorporating high-quality external knowledge, further revealing the value of knowledge provided in FineFake dataset.

\subsubsection{Fine-grained Classification Task}
Accuracy, macro recall, macro precision and macro f1-score are employed in this task.
Experimental results are presented in Table \ref{tab:binary-classification}.
Notably, there is significant degradation in the performance of all methods on fine-grained classification. This indicates that the current models are not sufficiently effective in identifying the underlying causes of fake news, suggesting substantial room for improvement in fine-grained classification. KEAN's superior performance may be due to the utilization of both images and knowledge, which enhances the model's capability to comprehend the underlying reasons for falsehoods.

\input{Table/Domain_Adaptation}
\subsubsection{Multi-domain Adaptation Task}
Here we evaluate the performance of models under the multi-domain adaption scenario by the three domain adaptation tasks defined in Section~\ref{section:3.2}.
Table~\ref{tab:domain-adaptation} presents the experimental results. Given significant disparities in the percentage of positive and negative samples across certain platforms, e.g. official news with social media, the weighted f1 value is adopted as the evaluation index for all experiments.

In topic domain adaptation task, the entertainment and conflict topic are selected as the test data, while utilizing data from the remaining four topics for model training. The results reveal notable variations in model performance across the different topics. Notably, all model's performance is substantially declined from the entertainment topic to the conflict topic, indicating that the conflict topic exhibits greater variance in data distribution. This finding aligns with our analysis of FineFake in Section~\ref{section:3.2}. KEAN achieves most SOTA metrics in this task, showcasing its robustness and adaptability across different topics. Specifically, its ability to leverage external knowledge and cross-domain learning enables it to bridge the gap of distribution shift between training and testing data more effectively.

In platform domain adaptation task, politics data is chosen due to its high proportion among the six topics and its significant attention in research~\cite{wang2017liar,roy2019deep,krstovski2022evons}. Notably, there is a significant performance drop in all models during platform domain adaptation task, which can be attributed to the label shift between source and target platforms, leading to a higher likelihood of misclassification.
KEAN achieves most SOTA in both metrics, particularly when source and target datasets exhibit significant label shift problems, e.g. Snopes to official news. This is attributed to our utilization of re-weighting methods, enabling our model to possess superior generalization capabilities when encountering label shift issues.

In dual domain adaptation task, models are trained with conflict topic data from the source platform and tested with politics topic data from the target platform. Results show a significant decrease in performance across all models on this task, as crossing two domains exacerbates both covariate shift and label shift problems. The pronounced domain shift poses a formidable challenge to the model's generalization capability. For instance, when encountering substantial label and convariate shift between two news environments (e.g. official conflict data with social media politics data), models lose ability to distinguish real from fake news. This indicates the benchmark role of FineFake in future research on fake news detection, offering enhanced generalization capability and practical value. KEAN still reaches SOTA in most task settings, further confirming its cross-domain capability to address convariate shift and label shift simultaneously. 
\begin{figure}[!t]
\begin{center}
\includegraphics[scale=1.0]{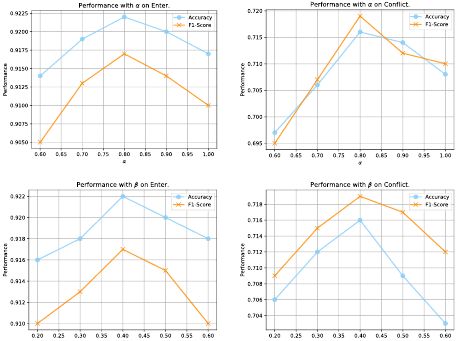}
\end{center}
\caption{Hyper-parameter sensitivity analysis of $\alpha$ and $\beta$.}
\label{fig:parameter}
\end{figure}
\subsection{Parameter Sensitive Analysis}
We perform a hyper-parameter sensitivity analysis on two key parameters: $\alpha$ and $\beta$ in Formula~\ref{loss}, denoting the weights of the adversarial loss $\mathcal{L}_{adv}$ and the reconstruction loss $\mathcal{L}_{recon}$, respectively. 
This analysis is conducted in the context of the topic adaptation task on social media platforms, with the results presented in Fig~\ref{fig:parameter}.
Notably, the model's performance under conflict topic is more susceptible to the values of $\alpha$ and $\beta$, possibly due to greater divergence between the data in the conflict topic and the training data.

When both $\alpha$ and $\beta$ are set too low, the model struggles to learn sufficient domain-invariant knowledge, which is critical for generalization across different topics. As a consequence, its performance in cross-topic fake news detection declines significantly. On the other hand, when $\alpha$ and $\beta$ are set too high, the model becomes overly focused on learning domain-invariant representations, which diminishes its ability to perform the primary task of fake news detection. This excessive emphasis on domain adaptation leads to a reduction in task-specific learning, ultimately degrading the model's overall performance. Balancing these two parameters is crucial to ensure the model captures both domain-invariant features and task-specific information effectively.

\subsection{Case Study}
\begin{figure}[!th]
    \setlength{\abovecaptionskip}{0.15cm}
    \begin{center}
    \includegraphics[scale=0.43]{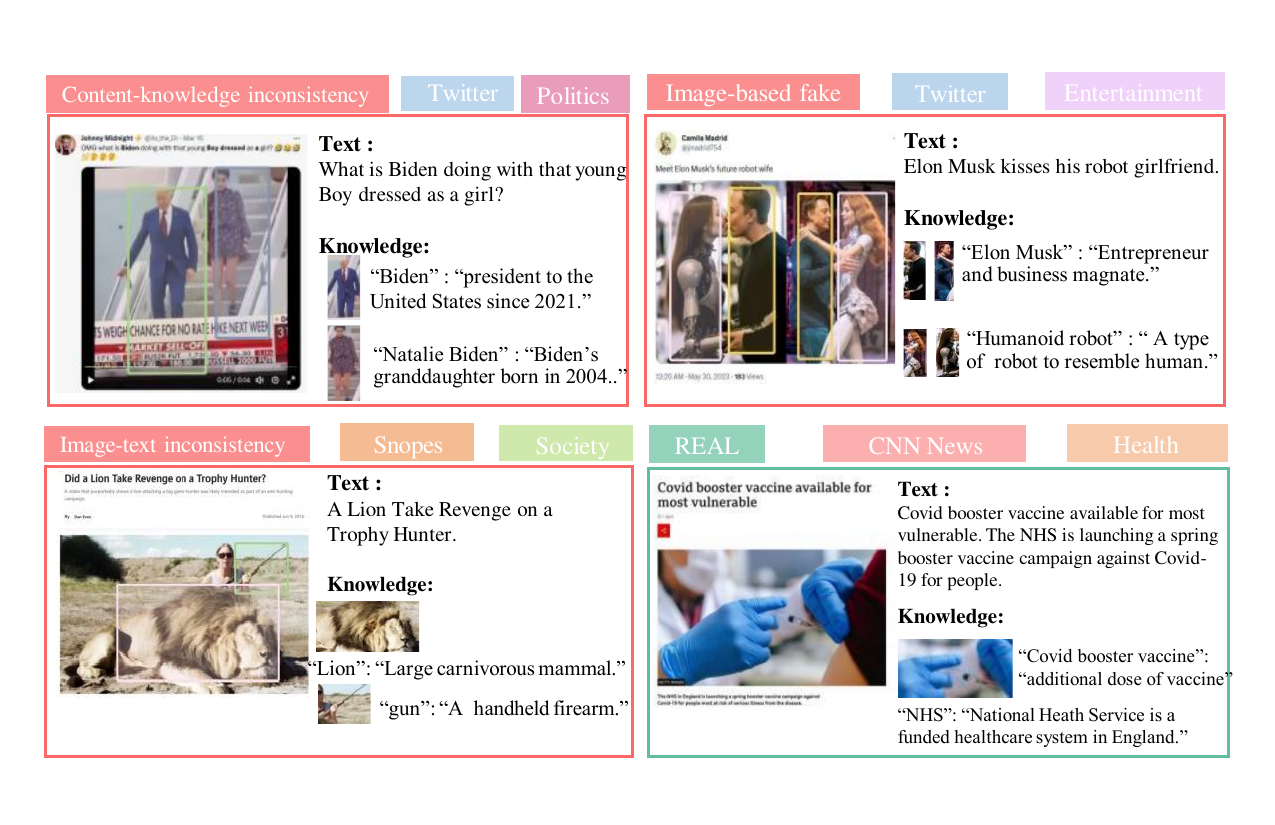}
    \end{center}
    \caption{Case studies on proposed dataset FineFake.}
    \label{fig:case_study}
\end{figure}
In this section, we provide four cases from four different topic domains and three platforms to provide an illustrative demonstration of the fine-grained annotations in Figure \ref{fig:case_study}. 
In the first case, the knowledge provides that the individual depicted in the picture is Biden's granddaughter instead of a boy. External knowledge is essential to refute this fake news, leading to content-knowledge inconsistency.
In the second case, the image is fabricated which makes its false reason image-based fake.  
In the third case,  the textual content claims the lion is taking revenge, but in the picture, the lion has already collapsed, leading to image-text inconsistency.  
The fourth case is a real news from CNN and thus there is no conflict between the image, text and external knowledge.

\section{Conclusion and Future Work}
\label{future work}
In this paper, we propose \textbf{FineFake}, a knowledge-enriched dataset for fine-grained multi-domain fake news detection. Each instance contains textual content, images, potential social connections, affiliated domain and other pertinent meta-data. FineFake empowers each news with rich and reliable external common knowledge and employs a fine-grained labeling scheme that classifies news articles into six distinct categories. 
We design three challenging tasks based on FineFake and conduct extensive experiments to provide reliable benchmarks for the community. Furthermore, we propose a \textbf{k}nowledge-\textbf{e}nhanced domain \textbf{a}daptation \textbf{n}etwork, dubbed \textbf{KEAN} for fake news detection to simultaneously solve convariate shift problem and label shift problem. 
Moving forward, our future research endeavors will involve expanding our modalities to include video data. 
\bibliographystyle{ieeetr}
\bibliography{sample}

\end{document}

%% file: Table/MVAE_ex_1.tex
    

\begin{table}[t]
  \centering
  \caption{Performance (accuracy) of MVAE under the cross-platform and cross-topic settings. The abbreviations of topics are: Pol: Politics; Ent.: Entertainment; Con.: Conflict.}
    \resizebox{1.0\linewidth}{!}{
    \begin{tabular}{c|rrr|c|rrr}
    \hline
    \toprule
    Task 1 & \multicolumn{3}{c|}{Cross-Platform } & Task 2 & \multicolumn{3}{c}{Cross-Topic} \\
    \midrule
    \multicolumn{1}{c|}{\diagbox[width=1.9cm]{Source}{ Target}} & \multicolumn{1}{l}{Red.} & \multicolumn{1}{l}{CNN} & \multicolumn{1}{l|}{Sno.} & \multicolumn{1}{c|}{\diagbox[width=1.9cm]{Source}{Target}} & \multicolumn{1}{l}{Pol.} & \multicolumn{1}{l}{Ent.} & \multicolumn{1}{c}{Con.} \\
    \midrule
    Reddit & \textbf{0.793} & 0.283 & 0.618 &  Pol. & \textbf{0.675} & 0.670  & 0.592 \\
    CNN & 0.327 & \textbf{0.810}  & 0.333 & Ent. & 0.614 & \textbf{0.742} & 0.646 \\
    Snope & 0.628 & 0.280  & \textbf{0.657} & Con. & 0.627 & 0.545 & \textbf{0.659} \\
    \bottomrule
    \hline
    \end{tabular}%
    }
  \label{mvae_ex_1}%
  \vspace{-3mm}
\end{table}%

%% file: Table/table_dataset.tex
\begin{table*}[!tbp]
  \centering
\renewcommand{\arraystretch}{0.9} 
\setlength\tabcolsep{4.5pt} 
\setlength{\abovecaptionskip}{0.1cm}
\setlength{\belowcaptionskip}{-0.05cm} 
  \caption{Comparison between FineFake and other fake news detection datasets.}
  \resizebox{0.99\linewidth}{!}{
    \begin{tabular}{l|ccc|cccc|c}
    \toprule
    \multirow{2}[4]{*}{} & \multicolumn{3}{c|}{Basic Info.} & \multicolumn{4}{c|}{Content Type} & \multicolumn{1}{c}{Annotation Info.} \\
\cmidrule{2-9}          & Size& Platform&Topic &Text&Image & Network&Knowl.&Label Type \\
    \midrule
    Breaking!\cite{pathak2019breaking} & 649 &BS Detector& US Election &\ding{51}&\ding{55}&\ding{55}&\ding{55}&Three Category \\
    Weibo21\cite{nan2021mdfend} & 9128 & Weibo &Nine Topics &\ding{51}&\ding{55} &\ding{55}&\ding{55}& Real/ Fake \\
    LIAR\cite{wang2017liar}  & 12,836 & Politifact & Political &\ding{51} &\ding{55}&\ding{55}&\ding{55}&Six Category \\
    Evons\cite{krstovski2022evons} & 92,969 &Media-source& US Election &\ding{51}&\ding{51}&\ding{55}&\ding{55}& Real/ Fake \\
    Weibo\cite{jin2017multimodal} & 9,528 & Weibo & ---    &\ding{51} &\ding{51} &\ding{55}&\ding{55}& Real/ Fake \\
    RD-E\cite{zubiaga2017exploiting} & 19,162 & Snopes/PolitiFact & ---      &\ding{51}&\ding{51}&\ding{55}&\ding{51}& Six Category \\
    Pheme\cite{zubiaga2017exploiting} & 5,802 & Twitter & News Events      &\ding{51}&\ding{51}&\ding{51}&\ding{55}& Real/ Fake \\
    FauxBuster\cite{zhang2018fauxbuster} & 917 & Twitter/Reddit & ---      &\ding{51}&\ding{51}&\ding{51}&\ding{55}& Real/ Fake \\
    Twitter\cite{boididou2015verifying} & 15,629 & Twitter & ---      &\ding{51}&\ding{51}&\ding{51}&\ding{55}& Real/ Fake \\
    MM-Covid\cite{li2020mm} & 11,173 & FullFact& Health &\ding{51}&\ding{51}&\ding{51}&\ding{55}& Real/ Fake \\
    MuMIN\cite{nielsen2022mumin} & 984 & Twitter &--- &\ding{51}&\ding{51} &\ding{51}&\ding{55}& Three Category \\
    MR$^2$\cite{hu2023mr2} & 14,700 & Twitter/Weibo &--- &\ding{51}&\ding{51} &\ding{51}&\ding{51}& Three Category \\  
    \midrule
    FineFake &16,909       &\makecell[c]{Snopes  \\ Social Media \\(e.g. Twitter) \\ Official News \\(e.g. CNN)}& \makecell[c]{Politics, Enter.\\Business, Health\\Society, Conflict}& \ding{51} & \ding{51} & \ding{51} & \ding{51} &  \makecell[c]{Real,Text/Image\\ Fake, Text-image/\\Content-knowledge\\ Inconsistency, \\ Others}\\
    \bottomrule
    \end{tabular}%
    }
  \label{tab:addlabel}%
  \vspace{-0.2cm}
\end{table*}%

%% file: Table/snopes_topic.tex
\begin{table}[t]
  \centering
  \caption{Mappings from the topic categories in Snopes to topics in FineFake.}
  \resizebox{1.0\linewidth}{!}{
    \begin{tabular}{ll}
    \toprule
    Topic & Snope Labels \\
    \midrule
    Politics & \makecell[l]{Ballot Box, Politicians, Politics, Race,\\ Racial Rumors, Soapbox, Conspiracy , Theorie,\\ Questionable Quotes,  Quotes} \\
    \midrule
    Enter. & \makecell[l]{Critter Country, Disney, Entertainment, \\ Holidays, Humor, Media Matters, Paranormal, \\ Social Media, Sports, Travel, Embarrassments} \\
    \midrule
    Business & \makecell[l]{Business, Charity, Gender Issues,  Immigration, \\ Law Enforcement, Legal, Legal Affairs, \\ Product Recalls, Risqué Business, Fraud\&Scams} \\
    \midrule
    Health & Abortion, Health, Medical \\
    \midrule
    Society & \makecell[l]{Automobiles, Climate Change , Cokelore,\\Computers, Education, Environment,  Rebellion,  \\Inboxer Rebellion,Love, Luck, Food, Science, \\ Sexuality, Technology, Weddings,\ Fauxtography, \\ Language, Junk New, Hurricane Katrina, College, \\ History, Glurge Gallery, Old Wives' Tales}  \\
    \midrule
    Conflict & \makecell[l]{Guns, Military, Viral Phenomena, Terrorism, \\ Horrors,  September 11th, Crime, Controversy} \\
    \bottomrule
    \end{tabular}%
    }
  \label{snope}%
\end{table}%

%% file: Table/dataset_analysis.tex
\begin{table}[t]
  \centering
  \renewcommand{\arraystretch}{0.99}
  \caption{The numbers of sample distributions, average words and average entities in different domains of FineFake.}
  \resizebox{1.0\linewidth}{!}{
    \begin{tabular}{l|c|c|c|c|c}
    \midrule
    Topic/Platform & \multicolumn{1}{l|}{Total} & \multicolumn{1}{l|}{Real} & \multicolumn{1}{l|}{Fake} & \multicolumn{1}{l|}{Words} & \multicolumn{1}{l}{Entities} \\
    \midrule
    Politics &5,727       &3,722       &2,005       &290.60       &3.00  \\
    Entertain. &3,699       &2,514       &1,185       &155.58       &2.33  \\
    Business &1,003       &527       &476       &308.10       &3.01  \\
    Health &710       &438       &272       &320.53       &3.01  \\
    Society &3,939       &2,236       &1,703       &133.94       &1.95  \\
    Conflict &1,718       &979       &739       &257.57       &2.62  \\
    \midrule
    Stream &5,000       &3,895       &1,105       &13.75       &1.25  \\
    Official &4,353       &4,138      &215       &813.02      &5.57  \\
    Fact-Check &7,556       &2,474       &5,082       &19.40       &1.73  \\
    \midrule
    The All &16,909       &10,507       &6,402       &222.03       &2.58  \\
    \bottomrule
    \end{tabular}%
    }
  \label{tab:dataset_analysis}%
\end{table}%

%% file: Table/binary_classification.tex
\begin{table*}[t]
  \centering
  \setlength{\abovecaptionskip}{1mm}
  \setlength{\belowcaptionskip}{-2mm}
  \caption{Classification task results, best results are in \textbf{bold} and second best results are \underline{underlined}.}
  \resizebox{1.0\linewidth}{!}{
    \begin{tabular}{cl|rrrr|rrrr|rrrr}
    \toprule
            \multirow{2}[4]{*}{Category} & \multicolumn{1}{c|}{\multirow{2}[4]{*}{Methods}}
           & \multicolumn{4}{c|}{Without Knowledge} & \multicolumn{4}{c|}{Knowledge Enhanced} & \multicolumn{4}{c}{ Fine-Grained Classification} \\
\cmidrule{3-14}          &       & \multicolumn{1}{l}{Acc} & \multicolumn{1}{l}{Pre} & \multicolumn{1}{l}{Recall} & \multicolumn{1}{l|}{F1} & \multicolumn{1}{l}{Acc} & \multicolumn{1}{l}{Pre} & \multicolumn{1}{l}{Recall} & \multicolumn{1}{l|}{F1} & \multicolumn{1}{l}{Acc} & \multicolumn{1}{l}{Pre} & \multicolumn{1}{l}{Recall} & \multicolumn{1}{l}{F1} \\
    \midrule
    \multirow{2}[2]{*}{\makecell[l]{\textbf{MultiModal} \\ \textbf{Method}}} & SAFE  &0.740       &0.751       &0.738       & 0.744      &  /     & /      & /      &  /  & 0.605      & 0.481      & 0.385      & 0.428\\
          & MVAE  &0.741       &0.738       & 0.728      &0.731       & /      &  /     & /      &/  &  0.550     & 0.448      & 0.315      & 0.307\\
    \midrule
    \multirow{3}[2]{*}{\makecell[l]{\textbf{Knowledge} \\ \textbf{Enhanced}}} & CompNet        & 0.780      &  0.779     &  0.767     &  0.772     & 0.791     & 0.794      & 0.779      & 0.786 
    & 0.656       & 0.605       & 0.483       &0.522\\
          & KAN   & 0.779      &  0.767     &  0.772     &  0.769     & 0.789     & 0.787      & 0.776      & 0.781 
    & 0.632       & 0.643       & 0.383 &0.424\\
          & KDCN & \underline{0.787}      &  0.783     &  \textbf{0.782}     &  0.782     & \underline{0.801}     & \underline{0.802}      & \textbf{0.791}      & \underline{0.796} 
    & 0.668       & 0.555       & 0.486 &0.499\\
    \midrule
    \multirow{3}[3]{*}{\makecell[c]{\textbf{Multi-Domain} \\ \textbf{Method}}} 
    & EANN & 0.785      &  0.789     &  0.774     &  0.781     & /     & /      & /      & / 
    & 0.644       & 0.665       & 0.431       &0.475 \\
    & MDFEND & 0.787       & 0.784      & \underline{0.781}      &  0.783     &  /     & /      & /      &  /  
    & 0.661      & \underline{0.676}      & 0.460      & 0.484\\
          & M3FEND & 0.783      & \textbf{0.793}      & 0.765      & 0.770     &  /     & /      & /      &  / 
          & \underline{0.680}      & 0.656      & \textbf{0.625}      & \underline{0.634}\\
          & CANMD &0.782       &0.768       & 0.773      & 0.770     &  /     & /      & /      &  /  
          & 0.676      & 0.637      & 0.597      & 0.602\\
    & KEAN & \textbf{0.790}      & \underline{0.791}      & 0.779      & \textbf{0.785}     &  \textbf{0.803}     & \textbf{0.806}      & \underline{0.788}     &  \textbf{0.797}
          & \textbf{0.692}      & \textbf{0.685}      & \underline{0.605}      & \textbf{0.638}\\
    \bottomrule
    \end{tabular}
    }
  \label{tab:binary-classification}%
\end{table*}%

%% file: Table/Domain_Adaptation.tex
\begin{table*}[t]
  \centering
  \setlength{\abovecaptionskip}{1mm}
  \setlength{\belowcaptionskip}{-0.5mm}
  \caption{Multi-domain task results, best results are in \textbf{bold} and second best results are \underline{underlined}. 
  In Task 1, we experiment with entertainment and conflict topics as tests, using the remaining four topics for training on the same platform.}
  \renewcommand{\arraystretch}{0.98}
  \resizebox{1.0\linewidth}{!}{
    \begin{tabular}{c|c|rrrr|rrrrrr|rrrrrr}
    \toprule
    \multirow{3}[6]{*}{Source } & Task  & \multicolumn{4}{c|}{Task 1 Topic DA} & \multicolumn{6}{c|}{Task 2 Platform DA}    & \multicolumn{6}{c}{Task 3 Dual DA} \\
\cmidrule{2-18}          & Target & \multicolumn{2}{c}{Enter.} & \multicolumn{2}{c|}{Conflict} & \multicolumn{2}{c}{Social Media} & \multicolumn{2}{c}{Official} & \multicolumn{2}{c|}{Snopes} & \multicolumn{2}{c}{Social Media} & \multicolumn{2}{c}{Official} & \multicolumn{2}{c}{Snopes} \\
\cmidrule{2-18}          & Metric & \multicolumn{1}{c}{Acc} & \multicolumn{1}{c}{F1} & \multicolumn{1}{c}{Acc} & \multicolumn{1}{c|}{F1} & \multicolumn{1}{c}{Acc} & \multicolumn{1}{c}{F1} & \multicolumn{1}{c}{Acc} & \multicolumn{1}{c}{F1} & \multicolumn{1}{c}{Acc} & \multicolumn{1}{c|}{F1} & \multicolumn{1}{c}{Acc} & \multicolumn{1}{c}{F1} & \multicolumn{1}{c}{Acc} & \multicolumn{1}{c}{F1} & \multicolumn{1}{c}{Acc} & \multicolumn{1}{c}{F1} \\
    \midrule
    \multirow{4}[2]{*}{\makecell[l]{Social \\ Media}}  & EANN &0.918      &0.915       & \underline{0.714}      &0.678       & /      & /      & 0.763      & 0.731      & 0.621      & 0.607      & /      & /      & 0.825      & 0.794      & 0.590      & 0.572  \\
    & MDFEND & 0.920      &\underline{0.919}      &0.697       &\underline{0.695}       & /      & /      & \underline{0.782}      & 0.727      & 0.615      & \underline{0.609}     & /      & /      & 0.802      & 0.789      & 0.697      & 0.681  \\
         & M3FEND &0.912       & 0.910      &0.649      &0.655      & /      & /      & 0.763      & 0.763      & \underline{0.631}      & 0.604      & /      & /      & \textbf{0.840}      & 0.804      & \underline{0.706}      & \underline{0.671}  \\
          & CANMD & \textbf{0.927}      &\textbf{0.923}       &0.697       &0.694       & /      &/       & \textbf{0.802}      & \textbf{0.810}      & 0.549      & 0.492      & /      &/       & 0.791      & \textbf{0.882}      & 0.653      & 0.434  \\
          & KEAN & \underline{0.922}      &0.917       &\textbf{0.716}       &\textbf{0.719}       & /      &/       & 0.776      & \underline{0.797}      & \textbf{0.634}      & \textbf{0.621}      & /      &/       & \underline{0.827}      & \underline{0.840}      & \textbf{0.710}      & \textbf{0.696}  \\
          
    \midrule
    \multirow{4}[2]{*}{Official}  & EANN &0.945       & 0.923      &0.838       & \underline{0.804}      &0.412       &0.427       &/       &/       &0.401       &0.379      &0.336       &0.232       &/       &/       &\underline{0.394}       &0.345  \\
    & MDFEND &\textbf{0.948}       &\underline{0.927}      & \underline{0.847}      &0.783       &\underline{0.679}       & \textbf{0.686}      & /      & /      & \underline{0.477}      & 0.479      &\underline{0.349}       & 0.378      & /      & /      & 0.334      & 0.169  \\
          & M3FEND &0.947       &0.921       &0.836       &0.798       & 0.669      & 0.537      & /      & /      & 0.411      & 0.383      &0.331      & 0.164      & /      & /      & 0.334      & 0.167  \\
          & CANMD & 0.947      & 0.921      &0.823       &0.782       & 0.487      & 0.425      & /      & /      & 0.351      & \underline{0.495}      & 0.331      & \textbf{0.497}      & /      & /      & 0.334      & \underline{0.473}  \\
          & KEAN & \underline{0.947}      & \textbf{0.968}      &\textbf{0.849}       &\textbf{0.828}       & \textbf{0.690}      & \underline{0.654}      & /      & /      & \textbf{0.493}      & \textbf{0.606}      & \textbf{0.402}      & \underline{0.480}      & /      & /      & \textbf{0.411}      & \textbf{0.495}  \\
    \midrule
    \multirow{4}[2]{*}{Snopes}  & EANN &0.653       &0.631       & \textbf{0.644}       & 0.610      & 0.721      & \underline{0.723}      & 0.774     & 0.718      & /      &/       & 0.686      & \textbf{0.695}      & 0.711    & \underline{0.714}      & /      &/  \\
    & MDFEND &\underline{0.661}       & \underline{0.647}      & 0.634      &\underline{0.628}       & 0.703      & 0.710      & \underline{0.784}      & 0.715      & /      & /      & 0.685      & 0.671      & \underline{0.737}      & 0.711      & /      & /  \\
          & M3FEND &0.632       &0.624       &0.615       &0.602       & \underline{0.721}      & 0.679      & 0.688      & 0.690      & /      & /      & 0.677      & 0.677      & 0.673      & 0.687      & /      & /  \\
          & CANMD & 0.644      & 0.563      & 0.590      & 0.538      & 0.669      & 0.491      & 0.678      & \textbf{0.801}      & /      & /     & 0.665      & 0.403      & 0.451      & 0.531      & /      & /  \\
          & KEAN & \textbf{0.663}      & \textbf{0.650}      & \underline{0.637}      & \textbf{0.635}     & \textbf{0.747}      & \textbf{0.730}      & \textbf{0.796}      & \underline{0.743}      & /      & /     & \textbf{0.688}      & \underline{0.679}      & \textbf{0.751}      & \textbf{0.724}      & /      & /  \\
    \bottomrule
    \end{tabular}%
    }
  \label{tab:domain-adaptation}%
  \vspace{-3mm}
\end{table*}%